\documentclass[runningheads]{llncs}
\usepackage{hyperref}
\usepackage{cite}
\usepackage{tabularx,booktabs}
\usepackage{bm}
\usepackage{enumitem}
\usepackage{amsfonts}
\usepackage{adjustbox}
\usepackage[misc,geometry]{ifsym}

\usepackage{algorithm} 
\usepackage{algpseudocode}

\hypersetup{
	colorlinks = true,
        urlcolor  = magenta,
	citecolor = blue,
	linkcolor = green,
	anchorcolor = black,
	filecolor = cyan, 
	menucolor = red, 
	runcolor = cyan,
	}

\makeatletter
\newcommand{\printfnsymbol}[1]{%
  \textsuperscript{\@fnsymbol{#1}}%
}
\makeatother
\usepackage{graphicx}
\begin{document}
\title{BoxShrink: From Bounding Boxes to Segmentation Masks}
%
%
\author{Michael Gr{\"o}ger\Letter\thanks{equal contribution}\orcidID{0000-0002-9732-7127}
\and
Vadim Borisov\printfnsymbol{1}\orcidID{0000-0002-4889-9989} 
\and
Gjergji Kasneci\orcidID{0000-0002-3123-7268}}
%
\authorrunning{M. Gr{\"o}ger et al.}
%
\institute{University of T{\"u}bingen, T{\"u}bingen, Germany \\
\email{michael.groeger@posteo.net}
}

\maketitle              

\begin{abstract}
One of the core challenges facing the medical image computing community is fast and efficient data sample labeling.
Obtaining fine-grained labels for segmentation is particularly demanding since it is expensive, time-consuming, and requires sophisticated tools. 
On the contrary, applying bounding boxes is fast and takes significantly less time than fine-grained labeling, but does not produce detailed results. 
In response, we propose a novel framework for weakly-supervised tasks with the rapid and robust transformation of bounding boxes into segmentation masks without training any machine learning model, coined BoxShrink. The proposed framework comes in two variants -- \textit{rapid}-BoxShrink for fast label transformations, and \textit{robust}-BoxShrink for more precise label transformations.   
An average of four percent improvement in IoU is found across several models when being trained using BoxShrink in a weakly-supervised setting, compared to using only bounding box annotations as inputs on a colonoscopy image data set.
We open-sourced the code for the proposed framework and published it online. 

\end{abstract}

\section{Introduction}
Convolutional neural networks (CNNs) have achieved remarkable results across image classification tasks of increasing complexity, from pure image classification to full panoptic segmentation, and have become, as a consequence, the standard method for these tasks in computer vision \cite{rawat2017deep}. However, there are also certain drawbacks associated with these methods. 
One of them is that in order to achieve satisfactory results, a data set of an appropriate size and high-quality labels are needed \cite{song2022learning}. The costs and time associated with labeling increase with the complexity of the task, with image classification being the cheapest and image segmentation being the most expensive one \cite{lin2014microsoft}. 
All of these challenges especially apply to medical artificial intelligence (MAI) applications since they depend on the input and feedback by expensive domain experts \cite{sourati2019intelligent}. 

In this work, we present a novel approach for fast segmentation label prepossessing, \textit{which is decoupled from any particular artificial neural network architecture}. The proposed algorithmic framework can serve as a first approach for practitioners to transform a data set with only bounding box annotations into a prelabeled (i.e., semantically segmented) version of the data set. Our framework consists of independent components such as superpixels \cite{stutz2018superpixels}, fully-connected conditional random fields \cite{krahenbuhl2013parameter} and embeddings. This makes it easy to add our framework to an existing machine learning pipeline.  

To evaluate the proposed framework, we select an endoscopic colonoscopy data set \cite{bernal2017comparative}. Multiple experiments show that our framework helps to considerably reduce the gap between the segmentation performance and efficiency of a neural network that is trained only on bounding boxes and one trained on fully segmented segmentation masks. 

The main contributions of this work are:
\begin{itemize}[noitemsep, topsep=0pt]
    \item We propose the BoxShrink framework consisting of two methods. One for a time-efficient and one for a more robust transformation of bounding-boxes into segmentation masks. In both methods there is no need to train a model.
    \item We publish our bounding-box labels for the CVC-Clinic data set for future research in the area of weakly-supervised learning.
    \item We open-source our code and publish it online.\footnote{\href{https://github.com/michaelgroeger/boxshrink}{https://github.com/michaelgroeger/boxshrink}}
\end{itemize}

\section{Related Work}
\label{sec:relatedwork}
In this Section, we further define weakly-supervised learning and separate it from other approaches such as semi-supervised learning. Also, we localize our work among those which use similar components. 

To reduce the need for resources such as time and money, various  learning methodologies were introduced such as semi-supervised and weakly-supervised learning~\cite{yang2021survey}. Semi-supervised learning leverages labeled data, e.g. for segmentation tasks correctly and fully segmented images and the availability of a larger amount of unlabeled data \cite{ouali2020, zou2020pseudoseg}. Weakly-supervised learning on the other hand, exploits noisy labels as a weak supervisory signal to generate segmentation masks. These labels can be provided in different forms such as points \cite{bearman2016s}, or image-level labels \cite{wang2021salient, kolesnikov2016seed}, being the more simpler ones, or more complex ones such as scribbles \cite{tang2018normalized, lin2016scribblesup}, or bounding boxes \cite{dai2015boxsup, khoreva2017simple}. A similar work \cite{xing2016weakly} to ours also utilizes superpixel embeddings and CRFs, but their method requires an additional construction of a graph of superpixels and a custom deep neural network architecture. Our method, on the other hand, is easier to integrate into existing pipelines. Also, in contrast to many other weakly-supervised approaches \cite{wei2017object, huang2018weakly}, we do not apply CRFs as a postprocessing step on the output of the model but as a preprocessing step on the input, hence, we leave the downstream model untouched. Furthermore, the proposed framework does not require special hardware such as GPU or TPU for the label preprocessing step. 

\begin{figure}[t]
	\centering
	\begin{minipage}{1.0\columnwidth}
		\centering
		\includegraphics[width=\textwidth, scale=0.48]{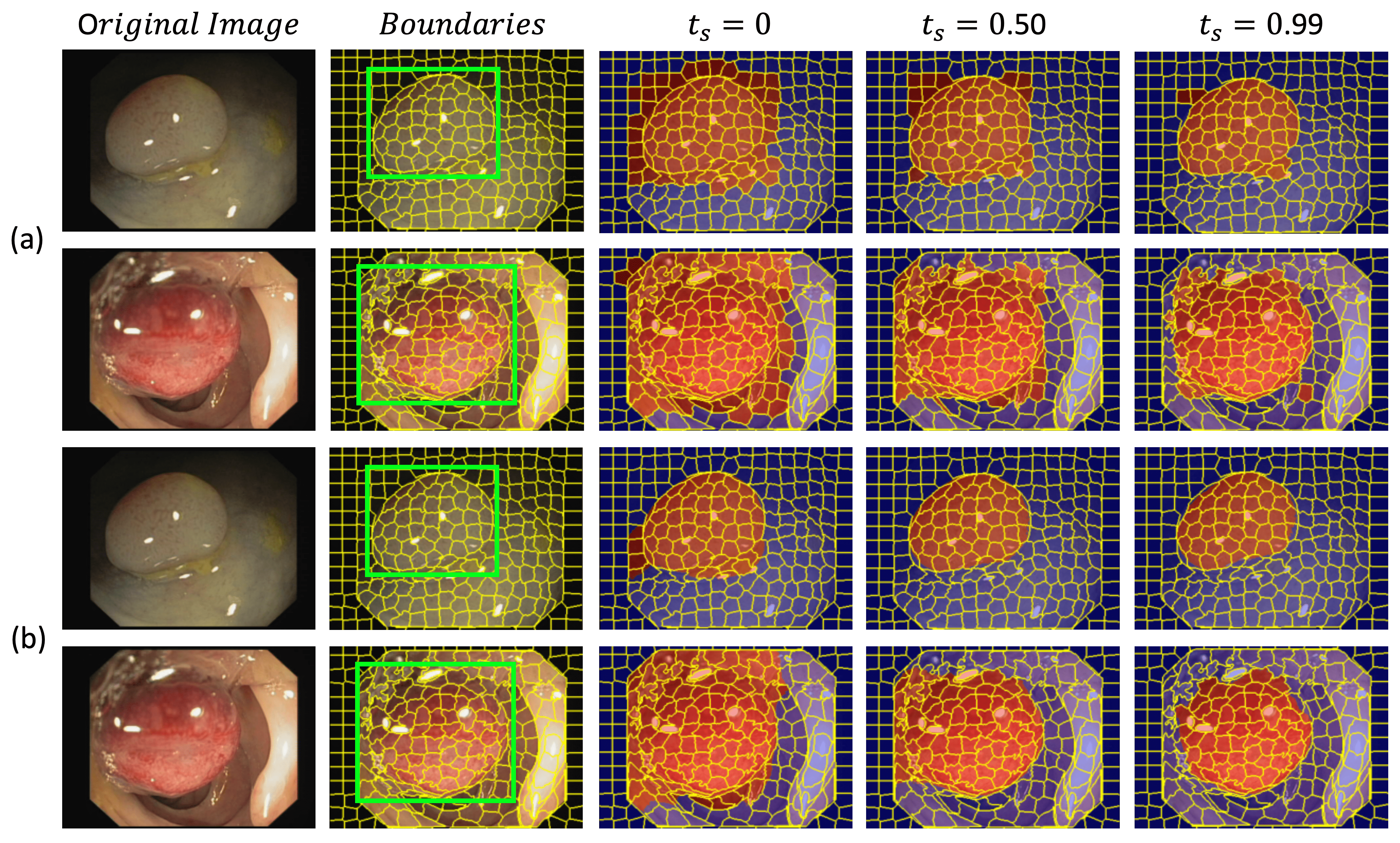}
		\caption{The impact of varying the threshold $t_s$, i.e., a hyperparameter of the BoxShrink framework for tuning the final segmentation quality, where (a) shows two data samples from the data set after the superpixel assignment step (Sec. \ref{sec:rapid_boxshrink}), and (b) demonstrates pseudo-masks after the FCRF postprocessing. As seen from this experiment, having a higher threshold might generate better masks but increases the risk of losing correct foreground pixels.}
		\label{fig:overlap_thresholding}
	\end{minipage}
\end{figure}

\section{Boxshrink Framework}
This section presents our proposed BoxShrink framework. First, we define its main components: superpixel segmentation, fully-connected conditional random fields, and the embedding step. We then explain two different settings of the framework, both having the same goal: to reduce the number of background pixels labeled as foreground contained in the bounding box mask.

\subsection{Main Components}
\textbf{Superpixels} aim to group pixels into bigger patches based on their color similarity or other characteristics \cite{stutz2018superpixels}. In our implementation, we utilize the SLIC algorithm proposed by \cite{achanta2012slic} which is a k-means-based algorithm grouping pixels based on their proximity in a 5D space. A crucial hyperparameter of SLIC is the number of segments to be generated which is a upper bound for the algorithm on how many superpixels should be returned for the given image. The relationship between the output of SLIC and the maximum number of segments can be seen in the supplementary material.
\label{superpixels}

\textbf{Fully-connected-CRFs} are an advanced version of conditional random fields (CRFs) which represent pixels as a graph structure. CRFs take into account a unary potential of each pixel and the dependency structure between that pixel and its neighboring ones using pairwise potentials \cite{triggs2007scene}. Fully-connected-CRFs (FCRFs) address some of the limitations of classic CRFs, such as the lack of capturing long-range dependencies by connecting all pixel pairs. Equation \ref{eq:FCRF} shows the main building block of FCRFs which is the Gibbs-Energy function \cite{krahenbuhl2011efficient}. 

\begin{equation}
        \label{eq:FCRF}
        E(x) = \sum_{i=1}^{N} \psi_{u}(x_{i}) + \sum_{i<j}^{N} \psi_{p}(x_{i}, x_{j}),
\end{equation}
where the first term $\psi_{u}(x_{i})$ measures the unary potential, that is, the cost if the assigned label disagrees with that of the initial classifier, the second term $\psi_{p}(x_{i}, x_{j})$ measures the pairwise potential, which is the cost if two similar pixels disagree on their label $x$. The input is over all pixels $N$. We use FCRFs to smooth the output pseudo-mask. 
\label{FCRF}

\textbf{Superpixel Embeddings} are a key component of the \textit{robust}-BoxShrink variant. The embedding function $M$ produces a numerical representation of every superpixel $k_i \in \bm K$ by returning an embedding vector. Formally, this operation can be depicted $M: \mathbb{R}^{m} \to \mathbb{R}^{n}$. 
Practically, this can be done by feeding each superpixel $k_i$ separately into a CNN model, such as a Resnet-50 \cite{he2016deep} pretrained on ImageNet \cite{deng2009imagenet}. By doing so, we obtain a 2048-dimensional vector representation for every superpixel. It allows us to get an aggregated representation of the foreground and background, by computing the mean embedding of all foreground and background superpixels in the training data set. These mean vectors are then used to assign superpixels either to the foreground or background class based on their cosine similarity.
\label{sec:embeddings} 

\subsection{\textit{rapid}-BoxShrink}
\label{sec:rapid_boxshrink}
We first split each image into superpixels using the SLIC algorithm for the \textit{rapid}-BoxShrink strategy. We overlap the superpixels with the provided bounding box mask and build a new mask based on those superpixels, which overlap the bounding box mask to a certain threshold. This approach is based on the assumption that the object of interest is always fully contained in the bounding box. The results depend on the number of segments generated which can be seen in the supplementary materials and the chosen threshold shown in Fig. \ref{fig:overlap_thresholding}. To this end, as shown in the supplementary material in Alg. 1, to make the final pseudo-mask more smooth, we run a FCRF as described in \ref{FCRF} on the thresholded superpixel mask.

\subsection{\textit{robust}-BoxShrink}
\label{sec:robust_boxshrink}

Leveraging the availability of superpixels, we also explore the use of embeddings to shrink the number of background pixels in the pseudo-mask. We segmented each image in the training data set into superpixels and then assigned them either to the foreground or background group by applying the thresholding approach as we have done it in the \textit{rapid}-BoxShrink variant (Sec. \ref{sec:rapid_boxshrink}). 
To generate the pseudo-masks, we start with the bounding box mask and segment the image using again the thresholding technique. This yields $\mathcal{F}$ superpixels for the foreground and $\mathcal{B}$ superpixels for the background. Then we go along the boundary foreground superpixels $\mathcal{F}_{o}$ and assign them either to the background or foreground class, depending on their cosine similarity score to the mean background and foreground embedding. The whole process can be seen in Fig. \ref{fig:embedding}.
The Alg. 2, which can be found in the supplementary materials, summarizes the main steps of the \textit{robust}-BoxShrink method.

\begin{figure}[t]
	\centering
	\begin{minipage}{1.0\columnwidth}
		\centering
		\includegraphics[width=\textwidth, scale=0.48]{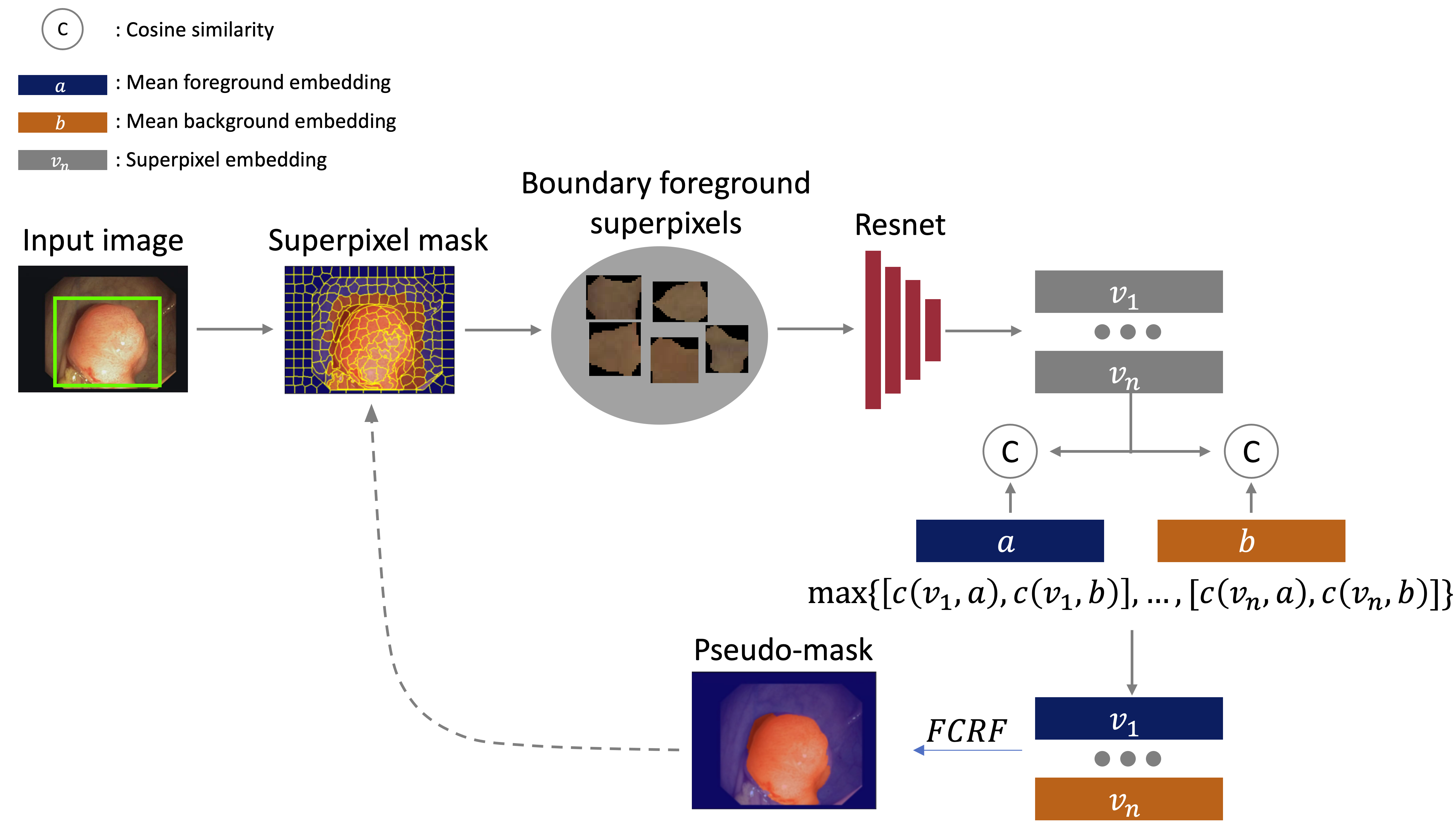}
		\caption{Overview of the \textit{robust}-BoxShrink method assuming the mean embedding vectors are given. First, we generate a superpixel mask based on the \textit{rapid}-BoxShrink approach but without utilizing the FCRF. Then, we extract each foreground superpixel on the boundary between foreground and background. Feeding each superpixel into a pretrained ResNet model yields one 2048-dimensional embedding vector per superpixel. Next, we calculate the cosine similarity score of each embedding and the mean background and foreground embedding. Based on the highest score we either keep the superpixel as foreground or assign it to the background class. Finally, we apply a FCRF on the resulting superpixel mask. The dashed line indicates that this approach can be run iteratively.}
		\label{fig:embedding}
	\end{minipage}
\end{figure}

\section{Experiments}
This Section presents qualitative and quantitative experiments for both versions of the BoxShrink framework. 

\textbf{Data set.} For all our experiments we utilize the endoscopic colonoscopy frames for polyp detection data set (CVC-Clinic DB)
\cite{bernal2017comparative}, it consists of 612 endoscopy images, each having a size of $288\times384\times3$. The data set comes along with binary ground truth segmentation masks, which we utilize for the evaluation of our weakly-supervised framework and to infer the bounding boxes. 
This data set was featured in multiple studies \cite{fan2020pranet, akbari2018polyp}.

\subsection{Qualitative and Quantitative Experiments}
\label{sec:qandqexp}
For our experiments, we utilize two popular deep learning architectures for segmentation tasks - U-Net~\cite{ronneberger2015u} and DeepLabV3+~\cite{chen2018encoder}. 

\textbf{Settings.} We have four settings, using: (1) Bounding boxes as labels which serves as our lower baseline, (2) labels generated with the \textit{rapid}-BoxShrink label transformation strategy, (3) labels generated with the \textit{robust}-BoxShrink label transformation strategy, and (4) a fully-supervised upper baseline with segmentation masks as labels. 

\textbf{Quality Measure.} We use the Intersection over Union (IoU) score as an evaluation measure. The IoU, also called Jaccard similarity $J$ between two sets $\bm A$ and $\bm B$, is a commonly used measure of how well the prediction aligns with the ground truth in image segmentation \cite{rahman2016optimizing}. As the equation below shows, the IoU is computed by dividing the intersection of two masks by their union. 

\begin{equation}
        \label{eq:Jaccard-Distance}
        J(\bm A, \bm B) = \frac{|{\bm A} \cap {\bm B}|}{|\bm A \cup \bm B|}.
\end{equation}

\begin{figure}[t]
	\centering
	\begin{minipage}{1.0\columnwidth}
		\centering
		\includegraphics[width=\textwidth, scale=0.48]{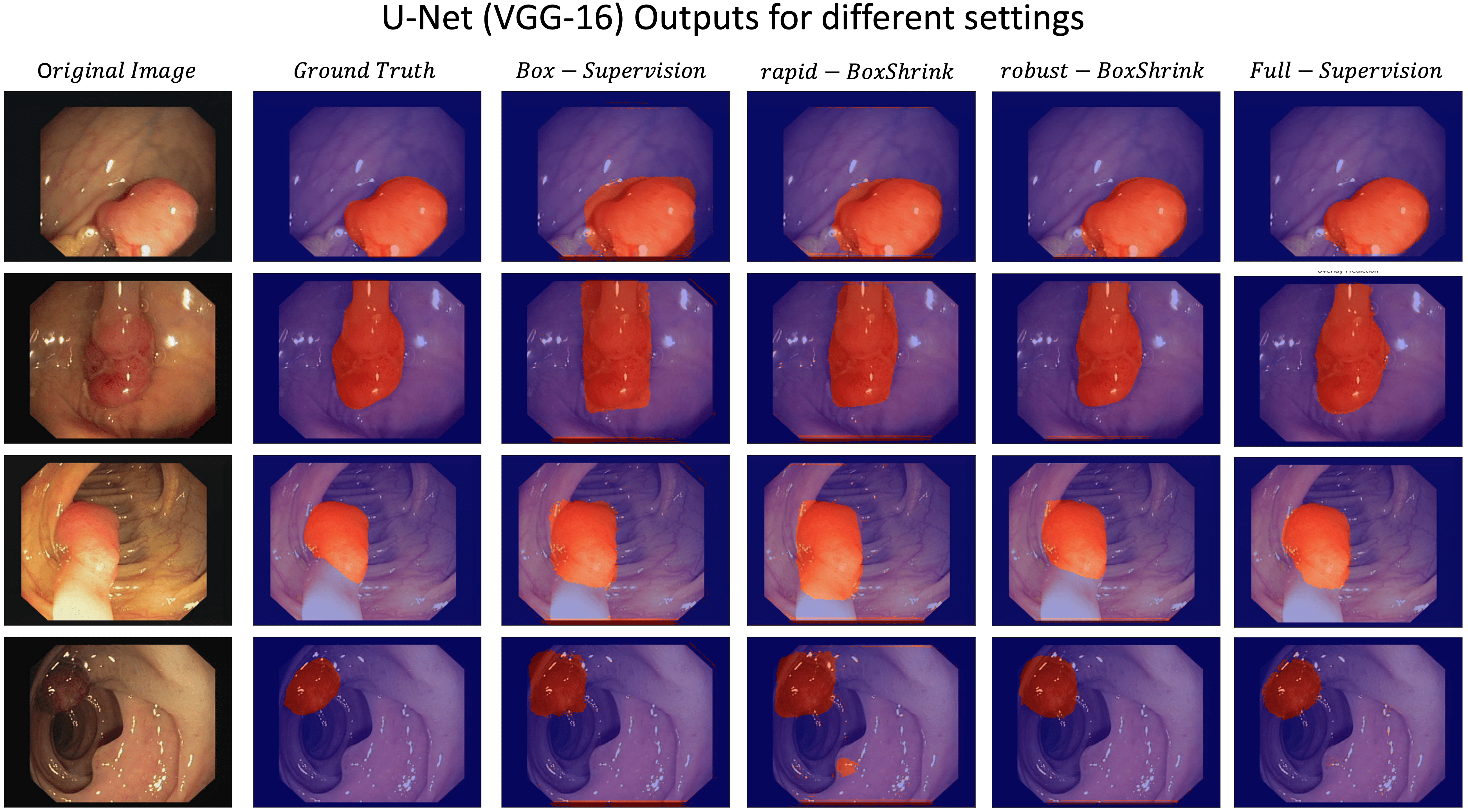}
		\caption{Qualitative model prediction masks on four random samples from the CVC-Clinic test set. The setting on which the model was trained on is indicated on top}
		\label{fig:qualitative_results}
	\end{minipage}
\end{figure}

\textbf{Results.} We present the quantitative results in Tab. \ref{tab:results}. 
In line with other publications, we also share situations where our presented Framework fails. Fig. 5, which can be found in the supplementary material, shows some examples. Fig. \ref{fig:qualitative_results} shows some good prediction masks from the test set made by models trained on the aforementioned four different settings.

\subsection{Reproducibility Details}
\label{sec:implementation}

We split the CVC-Clinic DB data set into 80 \% training data, 10 \% validation data and 10 \% test data. For splitting, we use the implementation from sklearn \cite{scikit-learn} with a random state of $1$. To generate the superpixel masks, we set the maximum number of segments $s$ to $200$, a threshold $t_s$ of $0.6$ for all training images and use the implementation from skimage \cite{van2014scikit}. To get the embeddings, we use a maximum number $s$ of 250 segments and a threshold $t_s$ of $0.1$ to not loose too much of the foreground. To smooth the superpixel masks we use the FCRF implementation provided by the pydensecrf package.\footnote{\href{https://github.com/lucasb-eyer/pydensecrf}{https://github.com/lucasb-eyer/pydensecrf}} 
Note that we do not train the FCRF (similar to\cite{huang2018weakly}) and set the FCRF hyperparmeters of the x/y-standard deviation for the pairwise Gaussian to $5$ and for the pairwise bilateral to $25$. We set the rgb-standard deviation to $10$. To determine the best performing model, we use the intersection over union (IoU) during training on the validation set. After the training, the best performing model is kept and evaluated once on the test set. Both, the test and validation set consist of ground truth masks. We generate all models using the segmentation-models PyTorch library.\footnote{\href{https://github.com/qubvel/segmentation_models.pytorch}{https://github.com/qubvel/segmentation\_models.pytorch}}

For our experiments we select ResNet-18, ResNet-50, and VGG-16 backbones pretrained on the ImageNet data set paired with U-Net and DeepLabV3+ as a decoder. We use the Sigmoid function as an activation function and the Adam \cite{kingma2014adam} optimizer with a learning rate of $0.0001$. As the loss function we utilize the Cross-Entropy Loss. 
During training, we apply step-wise learning rate scheduling where we decay the learning rate by $0.5$ each $5$ epochs. We train the ResNet-18 \& VGG-16 architecture for $25$ epochs and the ResNet-50 architecture for $15$ epochs. The training is being done on a 16 GB Nvidia Tesla P-100. We use a batch size of $64$ when using the ResNet-18, $32$ for the VGG-16 architecture and 16 when using ResNet-50. For both methods, \textit{rapid}-BoxShrink and \textit{robust}-BoxShrink, we return the initial bounding box mask if the total mask occupancy, that is the ratio of the bounding box and the total image is less than $0.1$ or the IoU between the pseudo mask and the bounding box mask is less than $0.1$. 
\begin{table}[t]
\centering
\caption{
Experimental results on the CVC-Clinic data set. All models are evaluated on the \textit{ground truth segmentation mask} in the validation and test set. 
The label format indicates the initial input label on which the model was either trained or our proposed frameworks were applied to.
The results are averages of six runs; we also report the corresponding standard deviation for each setting. This is being done to deliver a more consistent picture because of the random initialization of the decoder part and the stochasticity of the optimizer. The best performing results for our proposed methods are marked in bold. Higher IoU is better. 
}

\setlength{\tabcolsep}{12pt}
\adjustbox{max width=\textwidth}{
\begin{tabular}{lcccc} 
    \toprule
    Segmentation Model & Label Format & Backbone & Validation (IoU) & Test (IoU) \\
    \midrule
    U-Net & Bounding Boxes & VGG-16 & 0.749$\pm$0.023  & 0.772$\pm$0.030  \\
    U-Net (\textit{rapid}-BoxShrink) & Bounding Boxes & VGG-16 & 0.769$\pm$0.026  & 0.807$\pm$0.028  \\
    U-Net (\textit{robust}-BoxShrink)& Bounding Boxes & VGG-16 & \textbf{0.775}$\pm$0.013  & \textbf{0.824}$\pm$0.010  \\
    U-Net & Segment. Masks& VGG-16 & 0.796$\pm$0.025  & 0.829$\pm$0.025  \\
    \midrule
    U-Net & Bounding Boxes & ResNet-18 & 0.691$\pm$0.051  & 0.729$\pm$0.060  \\
    U-Net (\textit{rapid}-BoxShrink) & Bounding Boxes & ResNet-18 & 0.730$\pm$0.021 & 0.781$\pm$0.024 \\
    U-Net (\textit{robust}-BoxShrink)& Bounding Boxes & ResNet-18 & \textbf{0.755}$\pm$0.021 & \textbf{0.808}$\pm$0.021  \\
    U-Net &Segment. Masks & ResNet-18 & 0.800$\pm$0.032 & 0.859$\pm$0.044  \\
    \midrule
    U-Net & Bounding Boxes& ResNet-50 & 0.785$\pm$0.010  & 0.810$\pm$0.010  \\
    U-Net (\textit{rapid}-BoxShrink) &Bounding Boxes & ResNet-50 & 0.807$\pm$0.018  & 0.851$\pm$0.019 \\
    U-Net (\textit{robust}-BoxShrink) &Bounding Boxes & ResNet-50 & \textbf{0.813}$\pm$0.015  & \textbf{0.852}$\pm$0.012  \\
    U-Net & Segment. Masks & ResNet-50 & 0.889$\pm$0.012  & 0.920$\pm$0.016 \\
    \midrule
    DeepLabV3+ & Bounding Boxes & VGG-16 & 0.746$\pm$0.033  & 0.766$\pm$0.034  \\
    DeepLabV3+ (\textit{rapid}-BoxShrink) & Bounding Boxes & VGG-16 & \textbf{0.779}$\pm$0.023  & \textbf{0.817}$\pm$0.0201 \\
    DeepLabV3+ (\textit{robust}-BoxShrink) &Bounding Boxes & VGG-16 & 0.767$\pm$0.0187  & 0.809$\pm$0.024  \\
    DeepLabV3+ & Segment. Masks & VGG-16 & 0.832$\pm$0.049  & 0.858$\pm$0.051 \\
    \midrule
    DeepLabV3+ & Bounding Boxes & ResNet-18 & 0.723$\pm$0.025  & 0.758$\pm$0.021  \\
    DeepLabV3+ (\textit{rapid}-BoxShrink) & Bounding Boxes & ResNet-18 & 0.743$\pm$0.021 & 0.787$\pm$0.026 \\
    DeepLabV3+ (\textit{robust}-BoxShrink)& Bounding Boxes & ResNet-18 & \textbf{0.759}$\pm$0.005 & \textbf{0.806}$\pm$0.002  \\
    DeepLabV3+ &Segment. Masks & ResNet-18 & 0.808$\pm$0.010 & 0.844$\pm$0.012  \\
\end{tabular}
}

\label{tab:results}
\end{table}

\section{Discussion}
\label{sec:Discussion}

In this Section, we further discuss the application and future work of the proposed weakly-supervised framework. 

\textbf{The choice between \textit{rapid}-BoxShrink and \textit{robust}-BoxShrink}  depends on multiple factors - the time budget and expected label transformation quality. In our experiments, we observe that \textit{rapid}-BoxShrink takes on average only $0.5$ seconds to transform the labels for a singe data sample, where \textit{robust}-BoxShrink needs on average $3$ seconds to complete the label transformation, the processing time can be further optimized in future versions. However, from our extensive experiments (Section \ref{sec:qandqexp}), we can conclude that \textit{robust}-BoxShrink tends to outperform \textit{rapid}-BoxShrink in the weakly-supervised setting. The difference between the two variants is smaller for bigger models with \textit{rapid}-BoxShrink being once better than \textit{robust}-BoxShrink for the VGG-16 architecture. One explanation could be that bigger models are more robust to the label noise than smaller ones. We want to point out however, that the margin between the two is still overlapped by the standard deviations of both methods.

\textbf{Future work.} We want to further integrate the framework into the training pipeline by, e.g., adjusting the mean foreground and background embeddings as the model gets better. Also, we have evaluated our approach on a medium-sized data set with binary class segmentation. For a more detailed quality evaluation, an analysis of BoxShrink's performance on multi-class problems and bigger data sets is required. Lastly, starting with BoxShrink pseudo-masks instead of bounding box annotations directly could also improve existing state-of-the-art weakly-supervised learning algorithms.

\section{Conclusion}
\label{sec:Conclusion}

In this work, we presented BoxShrink, a weakly-supervised learning framework for segmentation tasks. 
We successfully demonstrate the effectiveness of the BoxShrink framework in the weakly-supervised setting on a colonoscopy medical image data set, where we employ bounding-box labeling and output the segmentation masks. Compared to the fully-supervised setting, our  weakly-supervised framework shows nearly the same results. Finally, we open-sourced and published the code and bounding boxes for the CVC-Clinic data set. 

%

\bibliographystyle{splncs04}
\bibliography{refs}


\appendix

\section{Supplementary Materials}

\begin{table}[H]
\centering
\setlength{\tabcolsep}{12pt}
\adjustbox{max width=\textwidth}{%

\begin{tabular}{lcccc} 
    \toprule
    Segmentation Model & Label Format & Backbone & Validation (LogLoss) & Test (LogLoss) \\
    \midrule
    U-Net & Bounding Boxes & VGG-16 & 0.508$\pm$0.026  & 0.512$\pm$0.020  \\
    U-Net (\textit{rapid}-BoxShrink) & Bounding Boxes & VGG-16 & \textbf{0.501}$\pm$0.038  & \textbf{0.505}$\pm$0.033 \\
    U-Net (\textit{robust}-BoxShrink) &Bounding Boxes & VGG-16 & 0.506$\pm$0.034  & 0.507$\pm$0.029 \\
    U-Net & Segment. Masks & VGG-16 & 0.520$\pm$0.032  & 0.518$\pm$0.031 \\
    \midrule
    U-Net & Bounding Boxes & ResNet-18 & 0.546$\pm$0.046  & 0.540$\pm$0.045  \\
    U-Net (\textit{rapid}-BoxShrink) & Bounding Boxes & ResNet-18 & 0.532$\pm$0.032 & 0.546$\pm$0.046 \\
    U-Net (\textit{robust}-BoxShrink)& Bounding Boxes & ResNet-18 & \textbf{0.501}$\pm$0.033 & \textbf{0.496}$\pm$0.034  \\
    U-Net &Segment. Masks & ResNet-18 & 0.496$\pm$0.018 & 0.491$\pm$0.019  \\
    \midrule
    U-Net & Bounding Boxes& ResNet-50 & 0.435$\pm$0.036  & 0.439$\pm$0.037  \\
    U-Net (\textit{rapid}-BoxShrink) &Bounding Boxes & ResNet-50 & \textbf{0.433}$\pm$0.066  & \textbf{0.441}$\pm$0.063 \\
    U-Net (\textit{robust}-BoxShrink) &Bounding Boxes & ResNet-50 & 0.458$\pm$0.063  & 0.465$\pm$0.057  \\
    U-Net & Segment. Masks & ResNet-50 & 0.413$\pm$0.030  & 0.413$\pm$0.030 \\
    \midrule
    DeepLabV3+ & Bounding Boxes & VGG-16 & 0.517$\pm$0.043  & 0.513$\pm$0.041  \\
    DeepLabV3+ (\textit{rapid}-BoxShrink) & Bounding Boxes & VGG-16 & 0.521$\pm$0.049  & 0.517$\pm$0.050 \\
    DeepLabV3+ (\textit{robust}-BoxShrink) & Bounding Boxes & VGG-16 & \textbf{0.501}$\pm$0.037  & \textbf{0.497}$\pm$0.036 \\
    DeepLabV3+ & Segment. Masks & VGG-16 & 0.505$\pm$0.048  & 0.503$\pm$0.047 \\
    \midrule
    DeepLabV3+ & Bounding Boxes & ResNet-18 & 0.519$\pm$0.050  & 0.522$\pm$0.049  \\
    DeepLabV3+ (\textit{rapid}-BoxShrink) & Bounding Boxes & ResNet-18 & \textbf{0.491}$\pm$0.028 & \textbf{0.497}$\pm$0.033 \\
    DeepLabV3+ (\textit{robust}-BoxShrink)& Bounding Boxes & ResNet-18 & 0.529$\pm$0.039 & 0.529$\pm$0.037  \\
    DeepLabV3+ &Segment. Masks & ResNet-18 & 0.502$\pm$0.037 & 0.501$\pm$0.035  \\
    
\end{tabular}
}
\caption{
Experimental results on the CVC-Clinic data set. All models are evaluated on the \textit{ground truth segmentation mask} in the validation and test set. The label format indicates the initial input label on which the model was either trained or our proposed frameworks were applied to. The results are averages of six runs; we also report the corresponding standard deviation for each setting. The best performing results for our proposed methods are marked in bold. Lower LogLoss is better. 
}
\label{tab:results_logloss}
\end{table}

\begin{algorithm}[h]
\caption{\textit{rapid}-BoxShrink}
\label{alg:rapid}
\begin{algorithmic}
\Require Data set with $N$ images and $U$ bounding box masks, superpixel overlay threshold $t_s$, maximum number of segments $s$
\For{$image, label=1,2,\ldots,N,U$} 
        \State ${\bm K}\gets SLIC(n, s)$ \Comment{Get $K$ superpixels}
        \State $\mathcal{\bm F,\bm B} \gets \{\}, \{\}$ \Comment{Foreground and Background superpixels}
        \For{$i=\{1, 2, 3, ..., |\bm K| \}$}
        \Comment{Iterate over $\bm K$ superpixels}
        \State ${\mathcal{\bm F,\bm B}}\gets Assign(u_i, k_i, t_s)$ \Comment{Foreground if overlap($u_i$, $k_i$) $>=$ $t_s$, background otherwise}
        \EndFor
        \State $M_{refined} \gets FCRF(F, n)$ \Comment{Get refined mask}
        \State \Return $M_{refined}$
\EndFor
\end{algorithmic}
\end{algorithm}

\begin{figure}[]
	\centering
	\begin{minipage}{0.96\columnwidth}
		\centering
		\includegraphics[width=\textwidth, scale=0.48]{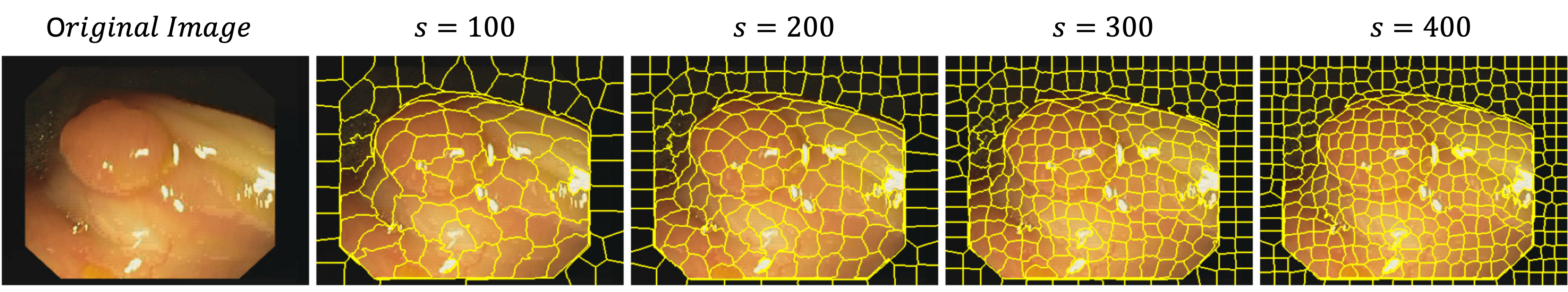}
		\caption{A demonstration of the SLIC algorithm~\cite{achanta2012slic} for different number of superpixels $s$ given a random sample from the CVC-Clinic data set}
		\label{fig:SLIC}
	\end{minipage}%
\end{figure}

\begin{algorithm}[]
\caption{\textit{robust}-BoxShrink}
\label{alg:robust_oslo}
\begin{algorithmic}
\Require Data set with $N$ images and $U$ bounding box masks or pseudo-masks, superpixel overlay threshold $t_s$, maximum number of segments $s$, Feature-Extractor $M$, Cosine-similarity function $C$, determine boundary superpixel function $O$, mean foreground embedding $\textbf{a}$ , mean background embedding $\textbf{b}$
\For{$image, label=1,2,\ldots,N,U$} 
        \State ${{\bm K} \gets SLIC(n, s)}$ \Comment{Get $|\bm K|$ superpixels using the SLIC method}
        \State $\mathcal{F,B} \gets \{\}, \{\}$ \Comment{Get foreground $\mathcal{F}$ and Background $\mathcal{B}$ superpixels}
        \For{$i=\{1, 2, 3, ..., |\bm K| \}$}
        \Comment{Iterate over $\bm K$ superpixels}
        \State ${\mathcal{F,B}\gets Assign(u_i, k_i, t_s)}$ 
        \EndFor
        \State ${\mathcal{F}_{o}\gets O(\mathcal{F,B})}$ \Comment{Determine boundary foreground superpixels}
        \For{$i=\{1, 2, 3, ..., |\mathcal{F}_o|\}$}
        \Comment{Iterate over $\mathcal{F}_o$ boundary foreground superpixels}
        \State ${\bm v_{i}\gets M(f_{oi})}$ \Comment{Get embedding for superpixel}
            \If{$C({\bm v_{i}},{\bm b}) > C({\bm v_{i}},{\bm a)}$}
            \State {$\mathcal{F}_{o} = \mathcal{F}_{o} \backslash {f_{oi}}$} \Comment{Remove a superpixel from foreground set}
		    \EndIf
		\EndFor
        \State $M_{refined} \gets FCRF(\mathcal{F}, n)$ \Comment{Get refined mask}
        \State \Return $M_{refined}$
        \EndFor
\end{algorithmic}
\end{algorithm}

\begin{figure}[]
	\centering
	\begin{minipage}{0.9\columnwidth}
		\centering
		\includegraphics[width=\textwidth, scale=0.48]{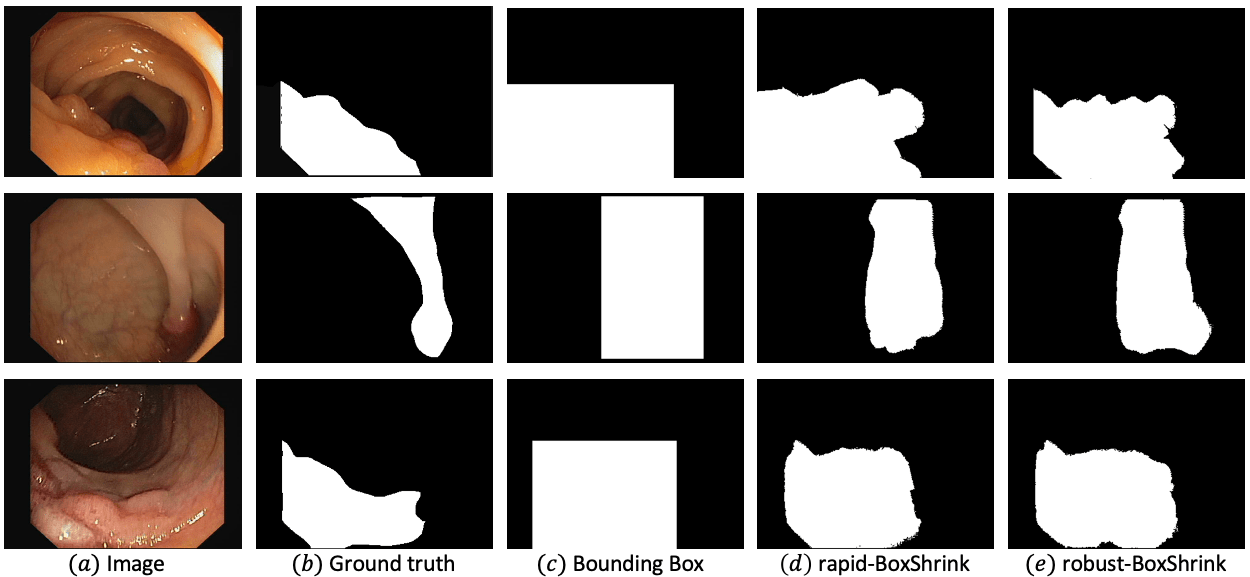}
	  	\caption{\textit{Failure cases} where both versions of the proposed segmentation framework do not eliminate enough background.}
		\label{fig:failure_cases}
	\end{minipage}%
\end{figure}
\label{results}

\end{document}